\documentclass{article}

\usepackage{microtype}
\usepackage{graphicx}
\usepackage{subfigure}
\usepackage{booktabs} 

\usepackage{hyperref}

\usepackage{amsfonts}
\usepackage{mathtools}
\mathtoolsset{showonlyrefs=true}

\usepackage{amsmath}
\usepackage[accepted]{icml2020}

\icmltitlerunning{Efficient sampling generation via Normalizing Flows}

\newcommand{\probtarget}[1]{p_{\text{target}}\left( #1 \right)}
\newcommand{\probp}[1]{p\left( #1 \right)}
\newcommand{\probq}[1]{q\left( #1 \right)}
\newcommand{\probnew}[1]{p_{\text{new}}\left( #1 \right)}
\newcommand{\sampleX}{\mathbf{x}}

\newcommand{\sampleU}{\mathbf{u}}
\newcommand{\probU}[1]{p_{\mathbf{u}}\left( #1 \right)}

\newcommand{\sampleZ}{\mathbf{z}}
\newcommand{\probqphi}[1]{q_\phi\left(#1 \right)}
\newcommand{\probsupport}[1]{p_{\text{support}}\left( #1 \right)}

\newcommand{\expectationOver}[2]{\mathbb{E}_{#1}\left[ #2 \right]}

\DeclareMathOperator*{\argmax}{arg\,max}
\DeclareMathOperator*{\argmin}{arg\,min}

\begin{document}

\twocolumn[
\icmltitle{Efficient sampling generation from explicit densities via Normalizing Flows}

\icmlsetsymbol{equal}{*}

\begin{icmlauthorlist}
\icmlauthor{Sebastian Pina-Otey}{ifae,aia}
\icmlauthor{Thorsten Lux}{ifae}
\icmlauthor{Federico S\'anchez}{unige}
\icmlauthor{Vicens Gaitan}{aia}
\end{icmlauthorlist}

\icmlaffiliation{ifae}{ Institut de F\'{i}sica d\'{}Altes Energies (IFAE) - Barcelona Institute of Science and Technology (BIST), Bellaterra (Barcelona), Spain}
\icmlaffiliation{aia}{Aplicaciones en Informática Avanzada (AIA), Sant Cugat del Vallès (Barcelona), Spain}
\icmlaffiliation{unige}{University of Geneva, Section de Physique, DPNC, Geneva, Switzerland}

\icmlcorrespondingauthor{Sebastian Pina-Otey}{spina@ifae.es}

\vskip 0.3in
]

\printAffiliationsAndNotice{}

\begin{abstract}
For many applications, such as computing the expected value of different magnitudes, sampling from a known probability density function, the target density, is crucial but challenging through the inverse transform. In these cases, rejection and importance sampling require suitable proposal densities, which can be evaluated and sampled from efficiently. We will present a  method based on normalizing flows, proposing a solution for the common problem of exploding reverse Kullback-Leibler divergence due to the target density having values of 0 in regions of the flow transformation.  The performance of the method will be demonstrated using a multi-mode complex density function.
\end{abstract}

\section{Introduction}

Generating samples from a \textit{target probability density} $\probtarget{\sampleX}$ to obtain data sets or compute expected values are fundamental tasks in many science and engineering disciplines. The generated Monte Carlo (MC) data is especially important when working within a high-dimensional space, since the integrals become intractable to be computed analytically. Additionally, when trying to sample in a high-dimensional space, numerical methods such as Markov Chain Monte Carlo (MCMC) have to be applied, since finding an explicit inverse transform to a simple-to-sample base distribution is usually impossible. Other methods, such as rejection and importance sampling, rely on defining a suitable \textit{proposal density function} $\probq{\sampleX}$ similar enough to the target density in order to obtain a good estimate for either sampling or computing expectations efficiently, as will be shown in Sec.~\ref{Sec:RejectionAndImportanceSampling}.

The proposal density $\probq{\sampleX}$ has to satisfy additionally two properties: 
\begin{enumerate}
    \item Sampling from it has to be fast and efficient.
    \item The proposal density of the samples has to be evaluable.
\end{enumerate}
Normalizing flows provide an expressive family of parametrized density functions $\probqphi{\sampleX}$ which satisfy precisely these conditions. As it will be described in Sec.~\ref{Sec:NormalizingFlows}, they consist in finding a differentiable and invertible transformation from a base distribution to a target distribution which allows for both of these tasks, hence being the perfect candidates to define a suitable proposal function. 

This approach of utilizing normalizing flows for importance sampling has been proposed previously by \cite{Mller2018NeuralIS}. We have added two modifications to obtain an optimal proposal function through normalizing flows, discussed in Sec.~\ref{Sec:Method}:
\begin{itemize}
    \item Usage of the reverse Kullback-Leibler (KL) divergence as the objective function.
    \item Presentation of a solution to the vanishing (exploding) gradient for the (reverse) Kullback-Leibler divergence via a convex combination of the target density and a support density. 
\end{itemize}
In Sec.~\ref{Sec:ToyProblem}, a toy problem is defined, which has a complicated vanishing and multi-mode density. We show that by using the reverse KL divergence with the convex combination of target $\probtarget{\sampleX}$ and support densities $\probsupport{\sampleX}$, we can train the normalizing flow to find a adequate proposal function to perform rejection and importance sampling.

\section{Background}

In this Section we will discuss the theoretical background of both the sampling methodologies and the proposal function framework. Sec.~\ref{Sec:RejectionAndImportanceSampling} describes how Rejection and Importance sampling work, and how their performances depend on the proposal function. In Sec.~\ref{Sec:NormalizingFlows}, the concept of Normalizing flow is introduced to find such suitable proposal function. In particular we will focus on a particular implementation of Normalizing flows, the Neural Spline Flows (NSF) \cite{Durkan2019NeuralSF}.

\subsection{Rejection and Importance sampling}
\label{Sec:RejectionAndImportanceSampling}

Very often one is in the situation that one has a complex probability density function, $\probp{\sampleX}$, which can be evaluated for any given $\sampleX$ and one would like to generate a data set following this distribution. However, it is too complex as that one can sample directly from it. 
A common approach for these cases is to find a simpler function, $\probq{\sampleX}$, the {\it proposal function}, from which one can easily draw samples.

If the aim is to calculate the expected value $\expectationOver{\sampleX \sim \probp{\sampleX}}{f(\sampleX )}$ of a function $f(\sampleX)$ with the $\sampleX$ drawn from $\probp{\sampleX}$ a suitable MC technique is {\it Importance Sampling}. 

Using the proposal function, $\expectationOver{\sampleX \sim \probp{\sampleX}}{f(\sampleX )}$ can be approximated by:
\begin{align}
\expectationOver{\sampleX \sim \probp{\sampleX}}{f(\sampleX)} &= \int \probp{\sampleX} f(\sampleX)  \;d\sampleX\\
&= \int \probq{\sampleX} \frac{\probp{\sampleX}}{\probq{\sampleX}} f(\sampleX)  \;d\sampleX\\
&= \expectationOver{\sampleX \sim \probq{\sampleX}}{f(\sampleX)\frac{\probp{\sampleX}}{\probq{\sampleX}}}\\
&\approx \frac{1}{N} \sum_{n=1}^N w_n f(\sampleX_n)  \text{ with } \sampleX_n \sim \probq{\sampleX}, \label{Eq:importance}
\end{align}
where the factors $w_n=\probp{\sampleX_n}/\probq{\sampleX_n}$ are known as the {\it importance weights}. The distribution of the weights indicates the efficiency of the importance sampling. For the ideal case of $\probq{\sampleX}$ being identical to $\probp{\sampleX}$ all $w_n$ will be equal 1. A broad width of the weight distribution on the other hand indicates that a larger number of samples is necessary to achieve the same precision for the expected value.

The variance of the distribution can be estimated by 
\begin{align}
    \sigma^2_q \approx \frac{1}{N-1} \sum_{n=1}^N \left( w_n f(\sampleX_n)  - \hat{\mu} \right)^2 \text{ with } \sampleX_n \sim \probq{\sampleX}, \label{Eq:importanceVariance}
\end{align}
where $\hat{\mu}$ is the approximated expected value obtained from Eq.~\eqref{Eq:importance}. Note that the larger the values of $w_n$ are, the higher the estimated variance is. The error on the estimated mean value is then
\begin{align}\label{Eq:errorMean}
    \sigma_{\hat{\mu}} = \frac{\sigma_q}{\sqrt{N}},\text{ with } N=\text{\# samples.}
\end{align}

Another MC technique for sampling is {\it Rejection Sampling} which has the advantage over importance sampling that it allows to produce samples which directly follow the distribution $\probp{\sampleX}$, a feature interesting for example for high energy physics MC generators.

In rejection sampling, the proposal distribution is used to create a comparison function, $k\probq{\sampleX}$, with $k$ being a constant factor, which has to satisfy that
\begin{equation}
    k\probq{\sampleX}\geq \probp{\sampleX}\;\; \forall \;\sampleX:\probp{\sampleX}>0. \label{Eq:rejection}
\end{equation}
The procedure is the following: First a sample $\sampleX$ is generated following $\probq{\sampleX}$. In a second step a random number, $u$, is generated uniformly in the range $[0,k\probq{\sampleX}]$, $u\sim \text{Unif}(0,k\probq{\sampleX})$. If $u$ fulfills the condition $u\leq \probp{\sampleX}$, the sample is accepted; otherwise it is rejected. The probability that a sample is accepted is proportional to: $p_{\text{acc}}\propto 1/k$, i.e., $k$ gives an intuition of the number of tries until we obtain an accepted sample.

Thus, for both sampling techniques it is crucial to find a $\probq{\sampleX}$ as similar as possible to $\probp{\sampleX}$ which at the same time fulfills the condition that $\probq{\sampleX} > 0$ for all $\sampleX$ which fulfill $\probp{\sampleX} > 0$. If this last condition is not satisfied, the methods fail to produce the desired result. In the following Sections we describe a method to achieve this.

\subsection{Normalizing flows}
\label{Sec:NormalizingFlows}

In the following a short introduction to the concept of normalizing flows will be shown, continuing with a concrete implementation, the Neural Spline Flows, and finishing with the objective function to train this kind of neural networks.

\subsubsection{General introduction}

\textit{Normalizing flows} are a mechanism of constructing flexible probability densities for continuous random variables. A comprehensive review on the topic can be found in \cite{Papamakarios2019NormalizingFF}, from which a brief summary will be shown in this Section on how they are defined, and how the parameters of the transformation are obtained. 

Consider a random variable $\sampleU$ defined over $\mathbb{R}^D$, with known probability density $\probU{\sampleU}$. A normalizing flow characterizes itself by a \textit{transformation} $T$ from this known density to another density $\probtarget{\sampleX}$ of a random variable $\sampleX$, the \textit{target density}, in the same space $\mathbb{R}^D$, via
\begin{align}
    \sampleX = T(\sampleU)\text{, with }\sampleU\sim\probU{\sampleU}.
\end{align}
The density $\probU{\sampleU}$ is known as \textit{base density}, and has to satisfy that it is easy to sample from and easy to evaluate (e.g., a multivariate normal, or a uniform in dimension $D$). The transformation $T$ has to be invertible, and both $T$ and $T^{-1}$ have to be differentiable, i.e., $T$ defines a diffeomorphism over $\mathbb{R}^D$.

This allows us to sample from $\probtarget{\sampleX}$ by sampling from $\probU{\sampleU}$ and applying the transformation. Additionally, we are able to evaluate the target density by evaluating the base density using the change of variables for density functions,
\begin{align}
    \probtarget{\sampleX} = \probU{\sampleU} |\det J_T(\sampleU)|^{-1} \text{ with } \sampleU = T^{-1}(\sampleX),
\end{align}
where the Jacobian $J_T(\sampleU)$ is a $D\times D$ matrix of the partial derivatives of the transormation $T$:
\begin{align}
    J_T(\sampleU) = \left[\begin{array}{ccc}
         \frac{\partial T_1}{\partial u_1} & \cdots & \frac{\partial T_1}{\partial u_D}  \\
         \vdots & \ddots & \vdots \\
         \frac{\partial T_D}{\partial u_1} & \cdots & \frac{\partial T_D}{\partial u_D}
    \end{array}\right].
\end{align}

The transformation $T$ in a normalizing flow is defined partially through a neural network with parameters $\phi$, as will be described below. If the transformation is flexible enough, the flow could be used to sample and evaluate any continuous density in $\mathbb{R}
^D$. In practice, however, the property that the composition of diffeomorphisms is a diffeomorphism is used, allowing to construct a complex transformation via composition of simpler transformations. Consider the transformation $T$ as a composition of simpler $T_k$ transformations:
\begin{align}
    T=T_K\circ \cdots \circ T_1.
\end{align}
Assuming $\sampleZ_0=\sampleU$ and $\sampleZ_K = \sampleX$, the forward evaluation and Jacobian are
\begin{align}
    \sampleZ_k &= T_k(\sampleZ_{k_1}),\; k=1:K,\\
    |J_T(\sampleU)| &= \left|\prod_{k=1}^K J_{T_k}(\sampleZ_{k-1})  \right|.
\end{align}
These two computations (plus their inverse) are the building block of a normalizing flow \cite{Rezende2015VariationalIW}. Hence, to make a transformation efficient, both operations have to be efficient. From now on forth, we will focus on a simple transformation $\sampleX=T(\sampleU)$, since constructing a flow from it is simply making the composition.

To define a transformation satisfying both efficiency properties, the transformation is broken down into autoregressive one-dimensional ones:
\begin{align}
x_i = \tau(u_i;\mathbf{h}_i)\text{ with } \mathbf{h}_i = c_i(u_{<i};\phi),
\end{align}
where $x_i$ is the $i$-th component of $\sampleX$ and $u_i$ the $i$-th of $\sampleU$.  $\tau$ is the \textit{transformer}, which is a one-dimensional diffeomorphism with respect to $u_i$ with parameters $\mathbf{h}_i$. $c_i$ is the $i$-th \textit{conditioner}, which depends on $u_{<i}$, i.e., the previous components of $\sampleU$, and $\phi$, the parameters of the neural network. The transformer is chosen to be a differentiable monotonic function, since then it satisfies the requirements to be a diffeomorphism. The transformer also satisfies that it makes the transformation easily computational in parallel and decomposing the transformation in one dimensional autoregressive transformers allows the computation of the Jacobian to be trivial, because of its triangular shape. To compute the parameter $\mathbf{h}_i$ of each transformer, one would need to process a neural network with input $\sampleX_{<i}$ for each component, a total of $D$ times. 

\textit{Masked autoregressive neural networks} \cite{Germain2015MADEMA} enable to compute all the conditional functions simultaneously in a single forward iteration of the neural network. This is done by masking out, with a binary matrix, the connections of the $\mathbf{h}_i$-th output with respect to all the components with index bigger or equal to $i$, $\geq i$, making it a function of the $<i$ components.

The transformer can be defined by any monotonic function, such as affine transformations \cite{Papamakarios2017MaskedAF}, monotonic neural networks \cite{Huang2018NeuralAF,Cao2019BlockNA,Wehenkel2019UnconstrainedMN}, sum-of-squares polynomials \cite{Jaini2019SumofSquaresPF} or monotonic splines \cite{Mller2018NeuralIS,Durkan2019CubicSplineF,Durkan2019NeuralSF}. In this work we will focus on a specific implementation of monotonic splines, the Neural Spline Flows.

\subsubsection{Neural Spline Flows}

In their work on \textit{Neural Spline Flows}, \cite{Durkan2019NeuralSF} advocate for utilizing monotonic rational-quadratic splines as transformers $\tau$, which are easily differentiable, more flexible than previous attempts of using polynomials for these transformations, since their Taylor-series expansion is infinite, and which are analytically invertible. 

The monotonic rational-quadratic transformation is defined by a quotient of two quadratic polynomials. In particular, the splines map the interval $[-B,B]$ to $[-B,B]$, and outside of it the identity function is considered. The splines are parametrized following \cite{Gregory1982PiecewiseRQ}, where $K$ different rational-quadratic functions are used, with boundaries set by the pair of coordinates $\{(x^{(k)},y^{(k)}\}_{k=0}^K$, known as knots of the spline and are the points where it passes through. Note that $(x^{(0)},y^{(0)}) = (-B,-B)$ and $(x^{(K)},y^{(K)}) = (B,B)$. Additionally, the $K-1$ intermediate derivative values of the nodes need to be provided, and have to be positive for the spline to be monotonic. At the boundary points, derivatives are set to 1 to match the identity function. 

Having this in mind, the conditioner given by the neural network outputs a vector $\mathbf{h}=[\mathbf{h}^w,\mathbf{h}^h,\mathbf{h}^d]$ of dimension $3K-1$ for the transformer $\tau$, $c_i(u_{<i};\phi)=\mathbf{h}_i$. $\mathbf{h}^w$ and $\mathbf{h}^h$ give the width and height between the $K+1$ knots, while $\mathbf{h}^d$ is the positive derivative at the intermediate knots. 

\subsubsection{Objective function}

Neural density estimators, such as the normalizing flows, are used for two main tasks:
\begin{enumerate}
    \item Estimate a density $\probtarget{\sampleX}$ through samples $\{\sampleX_1,\dots,\sampleX_N\}$ in order to be able to evaluate it.
    \item Estimate a density $\probtarget{\sampleX}$ through its density function in order to sample data from it.
\end{enumerate}

In both cases, the desire is to minimize the discrepancy between the approximated density of the neural network $q_\phi$ and the target density $p_\text{target}$. A common measure of discrepancy is given by the Kullback-Leibler (KL) divergence. Because of its asymmetry, it can be defined in two ways:
\begin{enumerate}
    \item The forward KL divergence:
    \begin{align}
    & D_{\text{KL}}\big(\probtarget{\sampleX}\| \probqphi{\sampleX} \big) \\
    & = \int \probtarget{\sampleX} \log \left( \frac{\probtarget{\sampleX}}{\probqphi{\sampleX}}\right) \;d\sampleX.
    \end{align}
    When minimizing it with respect to the parameters $\phi$ of the neural network, the objective function becomes:
    \begin{align}
    &\argmin_\phi  D_{\text{KL}}\big(\probtarget{\sampleX}\| \probqphi{\sampleX} \big)  \\  
    & = \argmin_\phi -\int \probtarget{\sampleX} \log  \probqphi{\sampleX} \;d\sampleX\\
    & = \argmax_\phi \int \probtarget{\sampleX} \log  \probqphi{\sampleX} \;d\sampleX\\
    & \approx \argmax_\phi \sum \log \probqphi{\sampleX} \text{ with } \sampleX \sim \probtarget{\sampleX}.
\end{align}
    \item The reverse KL divergence:
    \begin{align}
    & D_{\text{KL}}\big(\probqphi{\sampleX} \big\|\probtarget{\sampleX}) \\
    & = \int \probqphi{\sampleX} \log \left( \frac{\probqphi{\sampleX}}{\probtarget{\sampleX}}\right) \;d\sampleX.\label{Eq:forwardKL}
    \end{align}
    When minimizing, the objective function becomes:
    \begin{align}
        &\argmin_\phi  D_{\text{KL}}\big(\probqphi{\sampleX} \big\|\probtarget{\sampleX} \big) \\
        &\argmin_\phi \int \probqphi{\sampleX} \log \left( \frac{\probqphi{\sampleX}}{\probtarget{\sampleX}}\right) \;d\sampleX\\
        & \approx \argmin_\phi \sum \log \frac{\probqphi{\sampleX}}{\probtarget{\sampleX}} \text{ with } \sampleX \sim \probqphi{\sampleX}.\label{Eq:reverseKL}
    \end{align}
\end{enumerate}
When considering the nature of the problem being solved, the forward KL divergence helps to estimate the target density from samples, since it does not require to evaluate it. In the case of reverse KL divergence, sampling form the target density is not required, but being able to evaluate it is. The target density $\probtarget{\sampleX}$ does not need to be normalized. 

In the following Sections we will consider the reverse KL divergence as our objective function, since our goal is to find a good approximation to a given target density.

\section{Method}
\label{Sec:Method}

Consider an analytical/numerical target density $\probtarget{\sampleX}$ from which one wants to either sample from to create a data set or extract an expected value over it. As described in Sec.~\ref{Sec:RejectionAndImportanceSampling}, to perform either rejection or importance sampling, a suitable proposal density $q(\sampleX)$ is necessary. In the following, we use normalizing flows, implemented as NSF, to find such proposal function $\probqphi{\sampleX}$. 

When using the reverse KL divergence in Eq.~\eqref{Eq:reverseKL} for this task, however, depending on the support of $\probtarget{\sampleX}$, the objective function might explode if $\sampleX \sim \probqphi{\sampleX}$ such that $\probtarget{\sampleX}=0$. For densities with small support area such as the one of the toy problem in Sec.~\ref{Sec:ToyProblem}, this could hold for a major proportion of $\sampleX \sim \probqphi{\sampleX}$ when starting the training, making the update  of the neural network's parameters $\phi$ not ideal. 

We propose to redefine the target function as a convex combination of the target distribution $\probtarget{\sampleX}$ and a support density $\probsupport{\sampleX}$:
\begin{align}\label{Eq:convexcombination}
    \probnew{\sampleX} = (1-\alpha)\cdot \probtarget{\sampleX}+\alpha \cdot \probsupport{\sampleX},
\end{align}
with $\alpha\in(0,1)$. Effectively, background noise in form of the support density is added to the target density, with magnitude proportional to $\alpha$.

The support function should be positive over a domain which is a union of the supports of both the target density $\probtarget{\sampleX}$ and the initialized neural network density $\probqphi{\sampleX}$ at the beginning of the training. In the case of NSF, for instance, the initial neural network is mapped into a bounding box of $[-B,B]^D$, suggesting that a good support function might be a multivariate normal density with mean zero and covariance matrix $\mathbb{I}^D\cdot B/3$, where $\mathbb{I}^D$ is the identity matrix in dimension $D$. Experimentally this has given better result than using a uniform distribution in the bounding box, due to the gradient of the density being different than zero.

If $\alpha$ is small, $\probqphi{\sampleX} = \probnew{\sampleX}\simeq \probtarget{\sampleX}$, allowing us to use $\probqphi{\sampleX}$ as a proposal function for rejection and importance sampling. An important remark is that $\probqphi{\sampleX}$ does not have to fit $\probtarget{\sampleX}$ nor $\probnew{\sampleX}$ perfectly for these algorithms to sample/perform expected value computation accurately from the target function. Since $\probtarget{\sampleX}$ can be evaluated, it will correct the small discrepancies through the algorithms from Sec.~\ref{Sec:RejectionAndImportanceSampling}. The discrepancies come from adding background noise through the support density and from the imperfection of the training, but will be in practice small enough to produce an efficient sampling as will be shown experimentally in Sec.~\ref{Sec:ToyProblem}. $\probnew{\sampleX}$ is used only for training $\probqphi{\sampleX}$ while $\probtarget{\sampleX}$ is used for the sampling algorithms. This approach avoids the issue of $\probtarget{\sampleX}=0$ when computing the divergence during the training while introducing only a small inefficiency.

\begin{figure*}[!t]
\vskip 0.2in
\begin{center}
\centerline{\includegraphics[width=\textwidth]{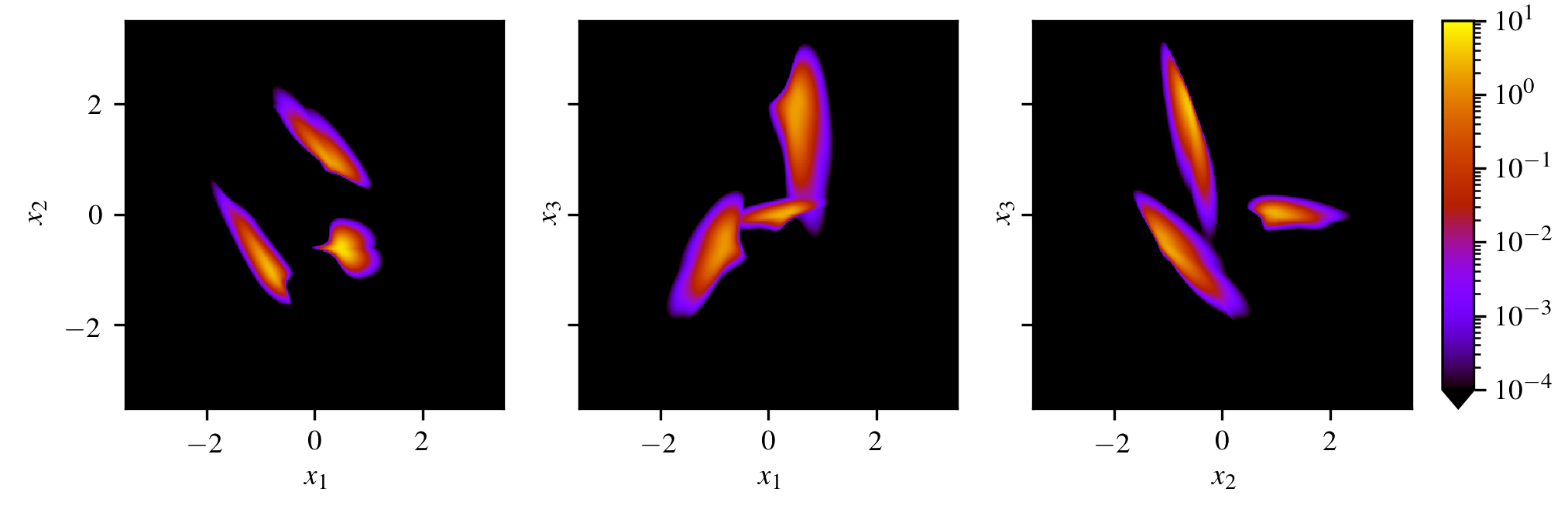}}
\caption{Three views of the toy model target density $\probtarget{\sampleX}$, with $\sampleX = (x_1,x_2,x_3)$, marginalizing one of the variables in each case to visualize the corresponding 2-dimensional densities. It displays the three sharp modes of the density, with a small support over the transformation area.}
\label{Fig:targetDensity}
\end{center}
\vskip -0.2in
\end{figure*}

The problem of exploding KL divergence is not unique to the reverse KL divergence. If instead the forward KL divergences, Eq.~\eqref{Eq:forwardKL}, is slightly modified using importance sampling to approximate the integral, as proposed in \cite{Mller2018NeuralIS}, the objective function has the expression:
\begin{align}
    \argmax_\phi \sum \frac{\probtarget{\sampleX}}{\probqphi{\sampleX}} \log \probqphi{\sampleX} \text{ with } \sampleX \sim \probqphi{\sampleX}.
\end{align}
Note that if $\sampleX \sim \probqphi{\sampleX}$ such that $\probtarget{\sampleX}=0$, then the neural network cannot update its parameters $\phi$ through training properly. The problem occurs under the exact same condition as when the reverse Kl divergence is used. The solution of tweaking the target density with a support function, Eq.~\eqref{Eq:convexcombination}, also helps in case one wants to use the forward KL divergence, as it eliminates this problem as well. 

\section{Toy Problem}
\label{Sec:ToyProblem}

A test of the previous described methodology was performed on a toy model. In Sec.~\ref{Sec:toyModel} the density is described as a combination of three modes of sharp densities. Then, the training procedure is detailed in Sec.~\ref{Sec:training} using a support density to avoid the exploding gradient of the reverse KL-divergence. The results\footnote{
Both training and results were performed on an Intel$^{\text{(R)}}$ Core$^{\text{(TM)}}$ i7-8700 @ 3.20GHz CPU with a GeForce RTX 2080Ti GPU machine.
}
of the training are discussed in Sec.~\ref{Sec:results}, where it is shown that the usage of a support function not only avoids the problem of the support domain, but also does not have a negative impact on the performance of the modified learned density.

\subsection{Toy multi-model description}
\label{Sec:toyModel}

The following toy model describes a non-trivial 3 dimensional density with 3 modes of data $\sampleX=(x_1,x_2,x_3)$. The three modes are transformations of a base density $p_\sampleU(\sampleU)$, following different rotations, scalings and translations in each of them (see Fig.~\ref{Fig:targetDensity}).

$\mathbf{u}\sim p_\mathbf{u}(\mathbf{u})$ is sampled through the following procedure:
\begin{align}
v_1 \sim & \;N(0,1),\\
(v_2+3) \sim & \;\text{Gamma}(|v_1|+3,0.3),\\
v_3 \sim & \; \text{SkewNormal}(|v_1\cdot v_2|)\\
\mathbf{v} =&\; (v_1,v_2,v_3),\\
R_\text{mix} =& \left( \begin{array}{ccc}
0.29 & -0.19 & 0.06 \\
-0.19 & 0.37 & 0.015 \\
0.06 & 0.015 & 0.11
\end{array} \right),\\
\mathbf{u} =&\; \mathbf{v} R_\text{mix},
\end{align}
with
\begin{itemize}
\item
$N(\mu,\sigma^2)$ the normal distribution.
\item
$\text{Gamma}(\alpha,\beta)$ the Gamma distribution with shape parameter $\alpha$ and scale parameter $\beta$.
\item
$\text{SkewNormal}(\alpha)$ is the Skew normal distribution with shape parameter $\alpha$.
\item
$R_\text{mix}$ an invertible matrix to mix the components, generated randomly once, fixed for purpose of reproducibility.
\end{itemize}

\begin{figure*}[!t]
\vskip 0.2in
\begin{center}
\centerline{\includegraphics[width=\textwidth]{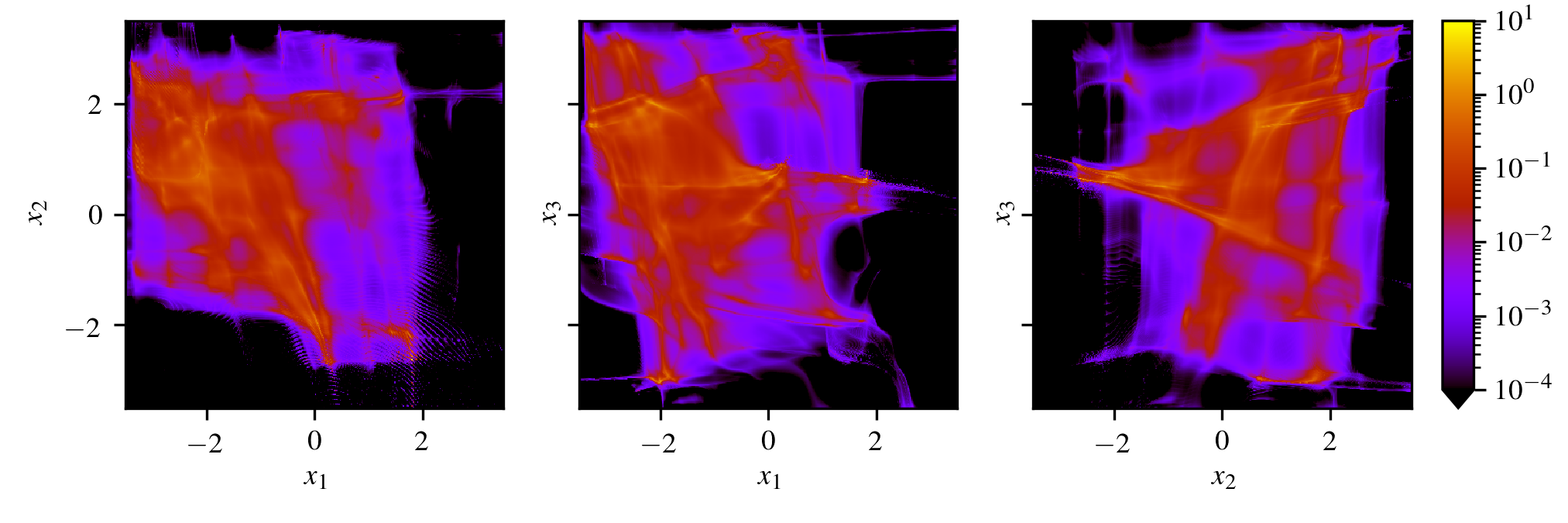}}
\caption{Three views of the marginalized initialized neural network density of the Neural Spline Flow. The density samples over the whole space initially, making highly likely that $\probtarget{\sampleX}=0$ with $\sampleX\sim\probqphi{\sampleX}$, thus producing the effect of exploding reverse KL-divergence of Eq.~\eqref{Eq:reverseKL}.}
\label{Fig:initialNN}
\end{center}
\vskip -0.2in
\end{figure*}

Starting from $p_\mathbf{u}(\mathbf{u})$, $\probtarget{\sampleX}$ is defined as
\begin{align}
\probtarget{\sampleX} &= \sum_{i=1}^3 \alpha_i p_\mathbf{u}(f_i(\mathbf{x})),\\
f_i(\mathbf{x}) &= (\mathbf{x}-\mathbf{t_i}) R(\theta_i)^{-1} \cdot s_i,\\
R(\theta) &= R_z(\theta)R_y(\theta)R_x(\theta),
\end{align}
with
\begin{itemize}
\item
$\alpha = (0.2,0.3,0.5)$ the weights of each mode.
\item
$\mathbf{t} = \big\{(0.0,1.5,0.0), (-1.2,-0.4,-0.9), (0.7,-0.6,$ $1.1)\big\}$ the translation vectors of each mode.
\item
$R(\theta)=R_z(\theta)R_y(\theta)R_x(\theta)$ a composition of rotations with angle $\theta$ on the three axes, with angles $\mathbf{\theta} = \big(0,\frac{2}{3}\pi,\frac{4}{3}\pi\big)$ for each mode.
\item
$s = (0.65,0.85,0.9)$ the scale factor of each mode.
\end{itemize} 

The three marginalized 2-dimensional densities can be seen in Fig.~\ref{Fig:targetDensity}, which are composed of three modes $p_\sampleU$, each scaled, rotated and translated according to the definition described above. This density has overall small support ($\approx 35\%$ of the $[-3.5,3.5]^3$ cube volume), centered in located regions of the volume.

\subsection{Training setup}
\label{Sec:training}

The setup for the Neural Spline Flow was the following (refer to \cite{Durkan2019NeuralSF} for a comprehensive description of the hyperparameters)\footnote{These hyperparameters were chosen due to the proximity of the nature of the density to the ones explored in the original NSF paper for similar problems and worked well.}: 128 hidden features, tail bound of 3.5 (i.e., the space transformed within the $[-3.5,3.5]^3$ cube), 10 composed transformations, 1024 batch size, validation size of 500k, $5\times10^{-5}$ learning rate, 100k training steps with cosine annealing scheduler. The complete set of configuration file can be found in the code attached. 

Fig.~\ref{Fig:initialNN} shows the initialized neural network density one could obtain, with the three 2-dimensional marginalized densities. Notice that the density spreads over the whole space in this particular case, making it likely to sample some point at a region outside of the support of the target density, thus causing the exploding reverse KL-divergence of Eq.~\eqref{Eq:reverseKL}. 

We redefine the target probability as described in Eq.~\eqref{Eq:convexcombination}, with a multivariate standard normal distribution in dimension 3 as the support function $\probsupport{\sampleX}$ and $\alpha=0.05$. With this new objective, the exploding divergence is no longer an issue, as $\probnew{\sampleX}>0$ in the $[-3.5,3.5]^3$ cube.

\begin{figure}[ht]
\vskip 0.2in
\begin{center}
\centerline{\includegraphics[width=\columnwidth]{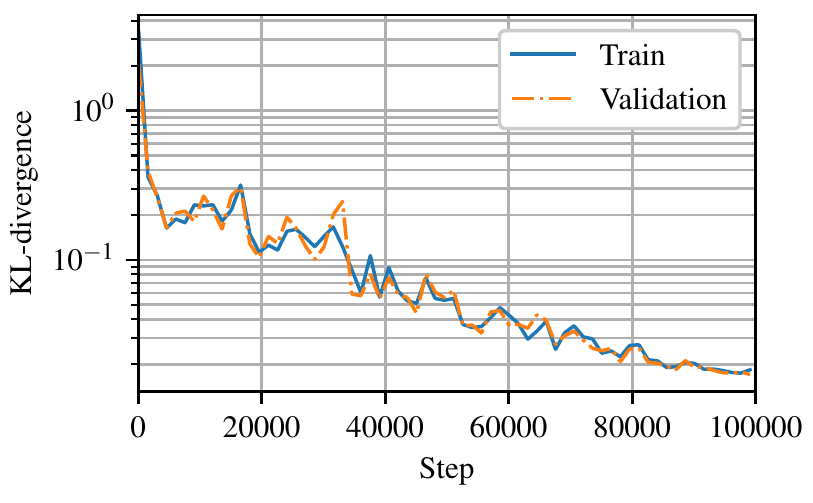}}
\caption{Kullback-Leibler divergence during the training of the modified density function with added support, converging to a value close to zero (0.017) for the training samples and the validation set. Training time was 5h 32m.}
\label{Fig:training}
\end{center}
\vskip -0.2in
\end{figure}

The training proceeds and converges to a value close to zero properly for the new modified target density, depicted in Fig.~\ref{Fig:training}. $\probqphi{\sampleX}$ is taken as the model with lowest KL-divergence for the validation set, with KL-divergence of value 0.017. Due to the nature of constantly generating \textit{new} samples during training from $\probqphi{\sampleX}$, no overfitting is possible.

\begin{table*}[!tp]
\caption{Weight statistical magnitudes for the three different proposal functions. The magnitudes are: mean, variance, maximum value, ratio of zeros, quantile .99 and quantile .9999. These statistics are computed over one million weights of each proposal function, where each weight is defined as $w(\sampleX)=\probtarget{\sampleX}/\probq{\sampleX}$ with $\sampleX \sim \probq{\sampleX}$ and $\probq{\sampleX}$ the proposal function.}
\label{Tab:weights}
\vskip 0.15in
\begin{center}
\begin{small}
\begin{sc}
\begin{tabular}{lrrrrrrrr}
\toprule
Proposal function &  $w_\text{mean}$ &    $w_\text{var}$ &    $w_\text{max}$ &  $w_\text{zeros}$ &  $w_{q_{.99}}$ &  $w_{q_{.9999}}$ \\
\midrule
Neural Spline Flow                 &   1.000 &    0.071 &   10.849 &    0.026 &  1.449 &    3.430 \\
Uniform             &   1.014 & 1942.702 & 5554.228 &    0.648 &  0.001 & 2343.291 \\
Multivariate Normal &   1.000 &  124.741 &  500.353 &    0.645 & 22.366 &  345.529 \\
\bottomrule
\end{tabular}
\end{sc}
\end{small}
\end{center}
\vskip -0.1in
\end{table*}

\subsection{Results}
\label{Sec:results}

For the purpose of comparison, two straightforward alternative proposal functions aside from the NSF where used:
\begin{itemize}
    \item A \textit{uniform distribution} over the cube $[-3.5,3.5]^3$.
    \item A \textit{multivariate normal distribution}. To find the appropriate mean and covariance matrix, one million samples were drawn uniformly in the cube $[-3.5,3.5]^3$, and assigned weights according to $\probtarget{\sampleX}$. The mean of the distribution is the weighted average and the covariance is the weighted covariance matrix multiplied by a factor 3 to cover a larger space.  
\end{itemize}

A sample size of one million is generated with each of the proposal, whose weights are computed as $w(\sampleX)=\probtarget{\sampleX}/\probq{\sampleX}$, with $\probq{\sampleX}$ either the NSF, the uniform or the multivariate normal distribution. In the Table~\ref{Tab:weights} statistics of the weights of these samples are shown. In particular the magnitudes are: mean, variance, maximum value, ratio of zeros, quantile .99 and quantile .9999. Notice that even though all of them have mean $\approx 1$ (although NSF has more significant digits), the variance and quantiles of the uniform and normal distributions are orders of magnitude bigger than the ones of the NSF. 

\begin{figure}[!ht]
\vskip 0.2in
\begin{center}
\centerline{\includegraphics[width=\columnwidth]{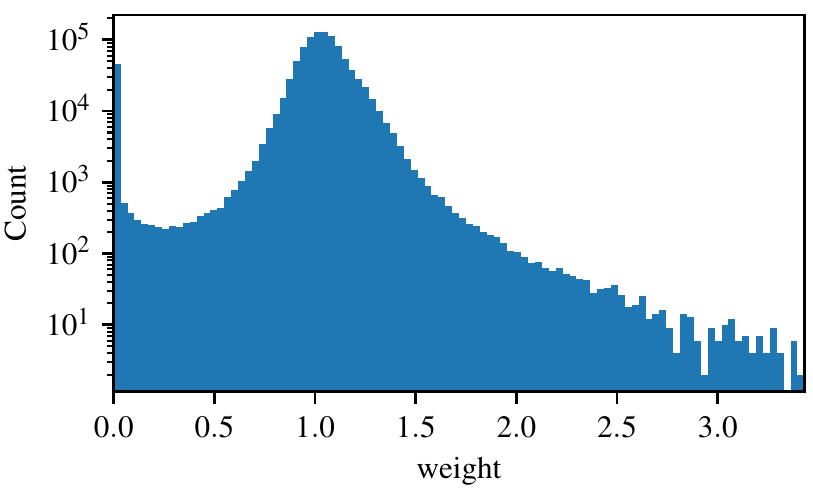}}
\caption{Weight distribution for the Neural Spline Flow proposal function. Weights are centered around 1, with a slight bias towards values greater than 1, due to the modified objective density through the convex combination.}
\label{Fig:weightsNN}
\end{center}
\vskip -0.2in
\end{figure}

\begin{figure}[!ht]
\vskip 0.2in
\begin{center}
\centerline{\includegraphics[width=\columnwidth]{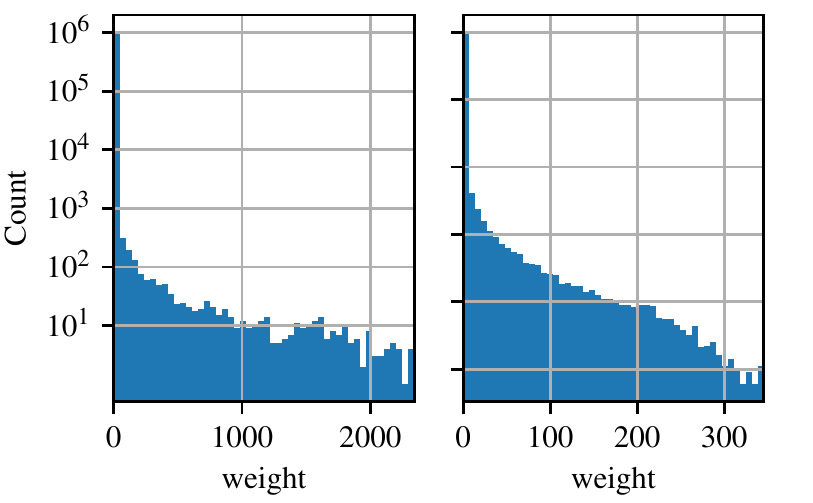}}
\caption{Weight values of the uniform (left) and multivariate normal (right) proposal functions. Long queues are observed, together with large values of the weights, aside from a great number of zero weight samples.}
\label{Fig:weightsUnifMN}
\end{center}
\vskip -0.2in
\end{figure}

Figures~\ref{Fig:weightsNN} and \ref{Fig:weightsUnifMN} show the explicit weight distributions. For the NSF case in Fig.~\ref{Fig:weightsNN}, notice how, aside from the 2.6\% of zeros, the weights are well distributed around 1, with a slight bias towards values bigger than 1. This is due to the convex combination of the target and the support function, making the new density to learn in general slightly smaller than the original one. In Fig.~\ref{Fig:weightsUnifMN}, long queues can be observed, which account for a bigger variance of the computed expectation value due to Eq.~\eqref{Eq:importanceVariance}. Additionally the proportion of zero weight values is quite large compared to the proportion of the NSF, as seen in Table~\ref{Tab:weights}.

To test the performance of the method, two functions of $\sampleX=(x_1,x_2,x_3)$ are defined whose expected values are computed with all three proposals:
\begin{align}
    f_1(\sampleX) &= x_1+x_2+x_3, \label{Eq:f1}\\
    f_2(\sampleX) &= \sqrt{x_1^2+x_2^2+x_3^2}. \label{Eq:f2}
\end{align}

First, the expected values over the target density for these two functions are computed using importance sampling with one million samples and their respective weights. Results are shown in Table~\ref{Tab:importance}, comparing them with the expectation with real samples of $\probtarget{\sampleX}$. Notice how for the NSF, the relative error of the mean is smaller, at least one order of magnitude better than the other two proposal functions. The error on the mean is of the same order of magnitude than the true one of the result.

\begin{table}[H]
\caption{Expected value of the functions $f_1(\sampleX)$ (Eq.~\eqref{Eq:f1}) and $f_2$ (Eq.~\eqref{Eq:f2}) computed using importance sampling over one million samples. The four results come from using exact samples, and the three proposals: Neural Spline Flow, Uniform and Multivariate Normal. NSF outperforms both other proposal by orders of magnitude both in mean and standard error values. The errors of the mean are computed using Eq.~\eqref{Eq:errorMean}.}
\label{Tab:importance}
\vskip 0.15in
\begin{center}
\begin{small}
\begin{sc}
\begin{tabular}{lrr}
\toprule
Proposal & $\mathbb{E}\left[f_1(\sampleX)\right]$ & $\mathbb{E}\left[f_2(\sampleX)\right]$\\
\midrule
Exact & $0.3963\pm0.0019$ & $1.6497\pm0.0004$\\
NSF & $0.3962\pm0.0019$ & $1.6502\pm0.0006$\\
Unif. & $0.4516\pm0.0801$ & $1.6546\pm0.0746$\\
M. N. & $0.4160\pm0.0201$ & $1.6562\pm0.0187$\\
\bottomrule
\end{tabular}
\end{sc}
\end{small}
\end{center}
\vskip -0.1in
\end{table}

Next, sets of one million samples from the target distribution are generated using rejection sampling, with a constant factor $k$ from Eq.~\eqref{Eq:rejection} equal to the quantile .9999 of the weights generated for Table~\ref{Tab:weights}. The results are shown in Table~\ref{Tab:rejection}, with the time it took to generate the data sets respectively with each different proposal function. Although all three proposal functions perform highly accurate regarding the results (as expected using rejection sampling), NSF performs 146 times faster than the uniform prior and 18.6 times faster than the multivariate normal one. Not only that, but taking into consideration that it took only 8.19s to sample one million samples, it can be used over importance sampling for this task, since the result is more accurate.

\begin{table}[H]
\caption{Expected value of the functions $f_1(\sampleX)$ (Eq.~\eqref{Eq:f1}) and $f_2$ (Eq.~\eqref{Eq:f2}) computed using rejection sampling over one million samples. The errors of the mean are computed using Eq.~\eqref{Eq:errorMean}. The four results come from using exact samples, and the three proposals: Neural Spline Flow, Uniform and Multivariate Normal. The results of all three proposals are similar quantitatively. The time (in seconds) however indicates a large difference from NSF with respect to the other two proposals, with a speedup of 146 times and 18.6 times respectively.}
\label{Tab:rejection}
\vskip 0.15in
\begin{center}
\begin{small}
\begin{sc}
\begin{tabular}{lrrr}
\toprule
Proposal & $\mathbb{E}\left[f_1(\sampleX)\right]$ & $\mathbb{E}\left[f_2(\sampleX)\right]$ & Time\\
\midrule
Exact & $0.3963\pm0.0019$ & $1.6497\pm0.0004$ & -\\
NSF & $0.4020\pm0.0019$ & $1.6508\pm0.0004$ & 8.26\\
Unif. & $0.3097\pm0.0019$ & $1.6438\pm0.0004$ & 1210\\
M. N. & $0.3939\pm0.0019$ & $1.6529\pm0.0004$ & 154\\
\bottomrule
\end{tabular}
\end{sc}
\end{small}
\end{center}
\vskip -0.1in
\end{table}

Overall, NSF provides an important improvement in the toy problem over the other two proposal functions for computing the expected value directly through importance sampling regarding precision (Table.~\ref{Tab:importance}) and time to sample exactly from the distribution through rejection sampling (Table.~\ref{Tab:rejection}).

\section{Future plans}

Normalizing flows offer a new and powerful way of finding adequate proposal functions for rejection and importance sampling in order to perform sampling and compute expected values. This has been seen in this work through the implementation of normalizing flows via neural spline flows, applying it to a complex multimode target density, with a small volume of non-zero values. The modification of the target density through a support function makes the tool useful for real world applications, which usually have narrow, delimited regions of non-zero density.

In the near future, the authors' aim is to apply this methodology tested here to real science problems. In particular the interest lies in applications to High Energy Physics to generate samples from a cross section and compute expectations related to it.

\bibliographystyle{icml2020}
\bibliography{bibliography}

\begin{thebibliography}{12}
\providecommand{\natexlab}[1]{#1}
\providecommand{\url}[1]{\texttt{#1}}
\expandafter\ifx\csname urlstyle\endcsname\relax
  \providecommand{\doi}[1]{doi: #1}\else
  \providecommand{\doi}{doi: \begingroup \urlstyle{rm}\Url}\fi

\bibitem[Cao et~al.(2019)Cao, Titov, and Aziz]{Cao2019BlockNA}
Cao, N.~D., Titov, I., and Aziz, W.
\newblock Block neural autoregressive flow.
\newblock In \emph{UAI}, 2019.

\bibitem[Durkan et~al.(2019{\natexlab{a}})Durkan, Bekasov, Murray, and
  Papamakarios]{Durkan2019CubicSplineF}
Durkan, C., Bekasov, A., Murray, I., and Papamakarios, G.
\newblock Cubic-spline flows.
\newblock \emph{ArXiv}, abs/1906.02145, 2019{\natexlab{a}}.

\bibitem[Durkan et~al.(2019{\natexlab{b}})Durkan, Bekasov, Murray, and
  Papamakarios]{Durkan2019NeuralSF}
Durkan, C., Bekasov, A., Murray, I., and Papamakarios, G.
\newblock Neural spline flows.
\newblock In \emph{NeurIPS}, 2019{\natexlab{b}}.

\bibitem[Germain et~al.(2015)Germain, Gregor, Murray, and
  Larochelle]{Germain2015MADEMA}
Germain, M., Gregor, K., Murray, I., and Larochelle, H.
\newblock Made: Masked autoencoder for distribution estimation.
\newblock In \emph{ICML}, 2015.

\bibitem[Gregory \& Delbourgo(1982)Gregory and
  Delbourgo]{Gregory1982PiecewiseRQ}
Gregory, J.~A. and Delbourgo, R.
\newblock Piecewise rational quadratic interpola-tion to monotonic data.
\newblock 1982.

\bibitem[Huang et~al.(2018)Huang, Krueger, Lacoste, and
  Courville]{Huang2018NeuralAF}
Huang, C.-W., Krueger, D., Lacoste, A., and Courville, A.~C.
\newblock Neural autoregressive flows.
\newblock \emph{ArXiv}, abs/1804.00779, 2018.

\bibitem[Jaini et~al.(2019)Jaini, Selby, and Yu]{Jaini2019SumofSquaresPF}
Jaini, P., Selby, K.~A., and Yu, Y.
\newblock Sum-of-squares polynomial flow.
\newblock In \emph{ICML}, 2019.

\bibitem[M{\"u}ller et~al.(2018)M{\"u}ller, McWilliams, Rousselle, Gross, and
  Nov{\'a}k]{Mller2018NeuralIS}
M{\"u}ller, T., McWilliams, B., Rousselle, F., Gross, M., and Nov{\'a}k, J.
\newblock Neural importance sampling.
\newblock \emph{ACM Trans. Graph.}, 38:\penalty0 145:1--145:19, 2018.

\bibitem[Papamakarios et~al.(2017)Papamakarios, Murray, and
  Pavlakou]{Papamakarios2017MaskedAF}
Papamakarios, G., Murray, I., and Pavlakou, T.
\newblock Masked autoregressive flow for density estimation.
\newblock In \emph{NIPS}, 2017.

\bibitem[Papamakarios et~al.(2019)Papamakarios, Nalisnick, Rezende, Mohamed,
  and Lakshminarayanan]{Papamakarios2019NormalizingFF}
Papamakarios, G., Nalisnick, E.~T., Rezende, D.~J., Mohamed, S., and
  Lakshminarayanan, B.
\newblock Normalizing flows for probabilistic modeling and inference.
\newblock \emph{ArXiv}, abs/1912.02762, 2019.

\bibitem[Rezende \& Mohamed(2015)Rezende and Mohamed]{Rezende2015VariationalIW}
Rezende, D.~J. and Mohamed, S.
\newblock Variational inference with normalizing flows.
\newblock \emph{ArXiv}, abs/1505.05770, 2015.

\bibitem[Wehenkel \& Louppe(2019)Wehenkel and
  Louppe]{Wehenkel2019UnconstrainedMN}
Wehenkel, A. and Louppe, G.
\newblock Unconstrained monotonic neural networks.
\newblock In \emph{BNAIC/BENELEARN}, 2019.

\end{thebibliography}

\end{document}